\pdfoutput=1
\documentclass{article}

\usepackage[nonatbib,preprint]{nips_2018}

\usepackage[english]{babel}

\usepackage[utf8]{inputenc} % allow utf-8 input
\usepackage[T1]{fontenc}    % use 8-bit T1 fonts
\usepackage{hyperref}       % hyperlinks
\usepackage{url}            % simple URL typesetting
\usepackage{booktabs}       % professional-quality tables
\usepackage{amsfonts}       % blackboard math symbols
\usepackage{nicefrac}       % compact symbols for 1/2, etc.
\usepackage{microtype}      % microtypography

\usepackage{amsmath,amssymb,amsthm}
\usepackage{algorithm2e}
\SetKwInOut{Parameter}{parameter}

\usepackage{graphicx}
\usepackage{caption}
\usepackage{subcaption}
\usepackage{wrapfig}

\DeclareMathOperator*{\argmin}{arg\,min}
\newcommand{\dd}{\text{d}}

\title{Deep Algorithms: designs for networks}
\author{
  Abhejit Rajagopal \\ %\thanks{Pre-print submitted to Neural Information Processing Symposium (NIPS) 2018.}
  Dept. of Electrical \& Computer Engineering\\
  University of California, Santa Barbara 93106\\
  \texttt{abhejit@ece.ucsb.edu} \\
   \And
   Shivkumar Chandrasekaran \\
   Dept. of Electrical \& Computer Engineering \\
   University of California, Santa Barbara 93106 \\
   \texttt{shiv@ece.ucsb.edu} \\
   \AND
    Hrushikesh Mhaskar \\
    Institute of Mathematical Sciences \\
    Claremont Graduate University \\
    \texttt{hrushikesh.mhaskar@cgu.edu}
}

\begin{document}

\maketitle

\begin{abstract}
    A new design methodology for neural networks that is guided by traditional algorithm design is presented. To prove our point, we present two heuristics and demonstrate an algorithmic technique for incorporating additional weights in their signal-flow graphs. We show that with training the performance of these networks can not only exceed the performance of the initial network, but can match the performance of more-traditional neural network architectures. A key feature of our approach is that these networks are initialized with parameters that provide a known performance threshold for the architecture on a given task.
\end{abstract}

\section{Introduction}
The original interest in neural networks arose from their connection with biology. Even though the input and output of biological networks are continuous, much of the current interest lies in static inputs and outputs. For such feed-forward networks there has existed a strong theoretical foundation based on approximation theory since the pioneering work of Cybenko, Mhaskar, and others \cite{cybenko1989approximation,mhaskar1992approximation,chui2016deep}. The recent resurgence of interest arises from the practical success of deep feed-forward networks on image classification problems, which in turn seems to be tied to several factors: the availability of massive training datasets, the availability of large amounts of cheap computing power, and the arrival of practical training algorithms \cite{rumelhart1986learning,taylor2010convolutional}. While the approximation theorists have shown the existence of good, but relatively large and shallow networks, practitioners have found success primarily with deep networks composed of several standard layers (convolutional, max-pooling, etc.) \cite{suzuki1998constructive,hornik1991approximation}. In particular, practitioners have primarily worked with a lego-block style approach to design, where they successively add and remove standard layers of varying tensor dimensions, until a sufficiently good design is arrived upon. Each iteration requires careful tweaking of the learning parameters and a carefully calibrated sense of when to pull the plug on a slowly converging network and try a new design. There are many papers devoted to this art with many case studies \cite{haykin2004comprehensive}. Recently this approach to network design has been ported to other non-classification problems \cite{kim2018communication}.

We propose a more systematic approach to the design of networks. In particular we claim that the design of a problem-specific (rather than data-specific) heuristic algorithm is key to the design of the network. We illustrate this design principle by several examples drawn from both classical classification problems and other less traditional machine learning areas.

Machine learning is normally used when the problem specification is so complicated as to defy a compact mathematical presentation. Typical in this area are image and video classification problems, where no precise mathematical formalism exists, but rather the problem is presented as a large corpus of ground-truth data. However, machine learning is also a valid approach when the mathematical problem is so difficult as to defeat the best effort of human algorithm designers to come up with a well-performing algorithm (both in terms of accuracy and speed). There are many classical problems that easily fall into the latter category. A simple example that immediately springs to mind are root finders for systems of polynomial equations. The literature on this area is classical and vast, and yet one can safely say that there exists no reliable practical algorithm that can be used in a black-box manner.
So, whether the problem is specified by data or mathematical formulas, one thing that is common is that the human algorithm designer has reason to believe that the problem is solvable and even has ideas on how to do so. We refer to these ideas as heuristics and will assume that they are presented as algorithms (or programs) that work reasonably well.

Our contention is that in many cases these heuristic algorithms can be viewed as special cases of a very large family of algorithms that can be parameterized by many real numbers. The initial heuristic itself can be viewed as a particular choice of these numerical parameters.

\textit{For instance}, when evaluating the similarity of two feature vectors $x,y \in \mathcal{R}^d$, it is natural to compare their distance in some norm. Usually in the absence of other information, the algorithm designer is likely to pick a familiar norm like the Euclidean distance:
\[
    d(x,y) = \|x-y\|_2.
\]
Being aware that this might not be the best choice, designers usually generalize to the Mahalanobis distance instead:
\[
    d^2_2(x,y) = (x-y)^T A (x-y)
\]
where $A\in\mathcal{R}^{d \times d}$~\cite{mignon2012pcca}. Notice, that the original distance measure is recovered for the choice $A=I_d$. We first observe that this is not the only such generalization. For example, one might embed this computation into an even larger computation graph, as:
\[
    d_k(x,y) = f\big( (I_k \otimes (x-y))^T A_k (I_k \otimes (x-y))^T \big)
\]
where the original distance could be recovered for some suitable choice of $f$ (e.g. average of the trace) and $A_k = I_k \otimes A$.

In particular note that the original heuristic distance is being recovered in every case by carefully selecting the numbers in a larger matrix. This corresponds to the insertion of additional edges in the first computational graph with trivial weights. One can generalize this observation in another direction too. For example by picking $A$ to be a Toeplitz matrix we get a \textit{convolutional} layer, and if we pick $A$ to be a Toeplitz-block-Toeplitz matrix we get a 2D convolutional layer. We can also choose $A$ to be the product of Toeplitz matrices in which case we would get several convolutional layers, and so on. Of course choosing $A$ to be a fully dense matrix would give us a full-connected layer at that stage of the computational graph.

% In particular, we notice that while these matrices are deliberately structured, many of the zero-entries can be thought of as additional parameters or connections in the computation graph that were previously not considered, in the same way that the signal-flow graph of neural networks are often constrained by architectural choices. Some simple examples in this case include picking the structure of $A$ as block-Toeplitz (corresponding to a CNN), or dense (corresponding to a fully-connected network).

We call this process of adding more weights as ``tensorization'' in general.

\vspace{-2mm}
\subsection*{Heuristics as trainable networks}
\vspace{-2mm}
We observe that when these heuristic networks are parametrized by real-numbers, they can be \textit{tuned} or \textit{calibrated} by special training algorithms. For example, if one wishes to use the currently popular deep neural network (DNN) training algorithms in TensorFlow or Pytorch, one could convert the heuristics (either by hand or special purpose compilers) into classical looking networks to achieve a guaranteed baseline performance, and then improve further by training.

To convert the heuristic into a network suitable for TensorFlow/PyTorch, we note that any finite sequence of code that only utilizes floating-point arithmetic operators is easily encoded as a classical network by just writing out its data-flow (signal-flow) graph.

\begin{itemize}
\item The first non-traditional construct would be \texttt{if-else} statements based on the truth values of numerical expressions involving inequality operators. In the network each of these truth values is encoded in a real weight as either a 1 or a 0. Then each piece of the heuristic guarded by \texttt{if-else} statements is split off into its own computational path in the network, and finally all the computational paths are added at the end of the \texttt{if-else} statement using the corresponding weights of the tests in guards. We just note that these can also be viewed as additional weights that will be tuned during the training phase.

\item The second non-traditional construct would be a \texttt{for} loop. If the number of loop executions is data independent then the \texttt{for} loop can be unrolled into a long sequence of statements. The number of layers that this generates in the network will be proportional to the number of loop executions. We conjecture that this is one of the primary reasons for the current crop of deep networks in classification problems.

\item If the \texttt{for} loop has a data dependent number of loop executions there are two possible strategies to follow. The simplest strategy is to choose a sufficiently large number of loop executions and just unroll the loop as before. In this case, care has to be taken to let the variables that are updated in the loop settle to their correct values by inserting suitable \texttt{if} statements. The second strategy would be to keep the data-dependent number of loop executions and use a simple adaptation in the DNN training algorithm instead. This strategy requires more space to explain and will be presented elsewhere.
\end{itemize}

% \subsection*{Heuristics as trainable networks}
%The main point here is that we can embed well-loved heuristics into a broader class of algorithms that are parameterised by many real variables \textit{with good starting values}. These heuristics can then be viewed as DNNs and trained by one's favorite ML framework. %with an established lower-bound on the best performance of a particular network architecture.

Now there are several questions that arise: Are there examples where this strategy actually shows improvement over the base heuristic? How does this strategy do in traditional image classification problems? When this strategy works, does it produce networks that do not look like current networks for the same problem? Is it possible to embed the heuristic network in a larger network that looks like the existing networks? Are good heuristics always deep or can they be shallow too?

% There are several questions that arise:
% \begin{itemize}
%     \vspace{-2mm}
%     \item Are there examples where this strategy actually shows improvement over the base heuristic?
%     \item How does this strategy do in traditional image classification problems?
%     \item When this strategy works, does it produce networks that do not look like current networks for the same problem?
%     \item Is it possible to embed the heuristic network in a larger network that looks like the existing networks?
%     \item Are good heuristics always deep or can they be shallow too?
% \end{itemize}

In this paper we present some early numerical evidence that purports to answer some of these questions.

\subsection{Related Work}
It is worthwhile at this point to note that similar ideas have definitely been expressed before outside the DNN literature. The classical ATLAS BLAS software, for example, would self-tune itself during installation by experimenting with a variety of integer parameters like block size \cite{demmel2005self}. Similarly the award winning FFTW code would self-tune several discrete parameters that determined which flavor of FFTs would be used at each stage of the recursion for different problem sizes \cite{frigo1998fftw}. However there was no emphasis on real parameters, and definitely no systematic use of a descent-based learning algorithm or a large corpus of ground-truth data.

Within the DNN literature itself, one can see a growing trend of composing classical convolutional layers, pooling layers, and fully connected layers, with some sort of meta-heuristics being used to justify the design \cite{grauman2011learning,han2015matchnet,pinheiro2016learning}. However, as far as we know, there has been no attempt to fully express how a heuristic leads precisely to a neural network (deep or otherwise).

% \newpage
\section{Designs for Networks} % /or/ Heuristics as Trainable Networks}
We now present two heuristics and demonstrate the incorporation of additional weights in their signal-flow graphs. We show that with training, the performance of these networks exceeds the performance of the initial network, and can also match the performance of more-traditional neural network architectures. We note that for improvement over the initial heuristic, it is sufficient that the gradient of the residual or error function with respect to the new parameters is non-zero, which is likely to be the case, and shows that human designers are very unlikely to produce optimal heuristics for a particular dataset or problem.
%% Please change this as necessary.

\subsection*{The Newton heuristic for solving systems of polynomial equations}

Our first example is a root-finder for a system of polynomial equations in $d$ variables:
\[
    \sum_{i \in R^d, \|i\|_1 < N} a_{j,i} x^i = 0, \qquad j = 1, \ldots, d,
\]
where $x\in\mathcal{R}^d$ is the unknown. It is well-known that this is an extremely difficult problem in floating-point arithmetic. One approach is to convert it into a polynomial equation in a single variable, but the price to pay is an exponential growth in the degree of the polynomial and potentially in the size of the coefficients. Another popular approach is to use a continuation technique \cite{allgower2012numerical}. However, the latter is notoriously difficult to implement well in floating-point arithmetic and slow to boot. Therefore the most popular approach is to use a locally convergent method like Newton, may be backed up with some kind of line search or back-stepping technique~\footnote{However it is difficult to get algorithms of this form to compute more than one root reliably as they need some kind of deflation technique to mask already computed roots.}. As one can see, all of these methods fall into the category of what we call heuristics.

If we represent the polynomial system as $F(x)=0$, then the simple Newton \textit{heuristic} could be expressed as follows:

\begin{minipage}[c]{.4\textwidth}
\begin{algorithm}[H]
    % \vspace{-3mm}
    $x_0 = 0$ \\
    \While{$\|F(x_{n})\| \leq \epsilon$}{
    $x_{n+1} = x_n + \alpha (F'(x_n))^{-1}F(x_n)$
    }
    % \captionof{algocf}{Newton heursitc.}\label{algo:NewtonHeuristic}
\end{algorithm}
% \vspace{-2mm}
\end{minipage}%
\begin{minipage}[c]{.6\textwidth}
    \centering
%   \rule{0.3\textwidth}{50pt}
    \includegraphics[width=\linewidth]{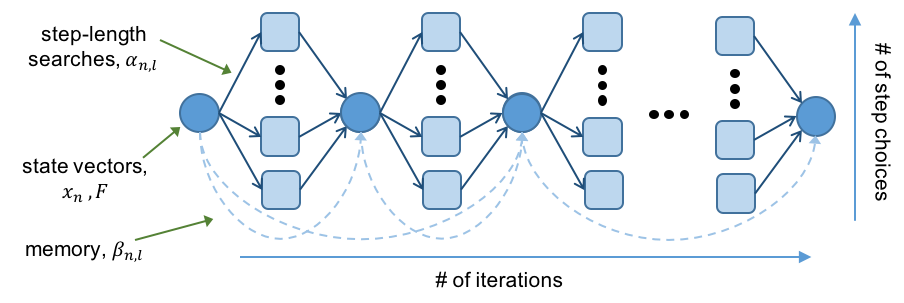}
    \captionof{figure}{The unrolled graph of Newton's method with line search resembles a DNN with pooling.} \label{fig:newton_graph}
%   \captionof{figure}{This is a figure caption.} \label{myfig1}
%   \bigskip
%   \rule{0.25\textwidth}{70pt}
%   \captionof{figure}{This is another figure.} \label{myfig2}
\end{minipage}

% \begin{algorithm}
%     $x_0 = 0$ \\
%     \While{$\|F(x_{n})\| \leq \epsilon$}{
%     $x_{n+1} = x_n + \alpha (F'(x_n))^{-1}F(x_n)$
%     }
%     %\caption{Newton heursitc.}\label{algo:NewtonHeuristic}
% \end{algorithm}
% \begin{figure}[hbt!]
%     \centering
%     \includegraphics[width=0.65\textwidth]{fig_newton_graph.png}
%     \caption{The unrolled graph of Newton's method with line search resembles a DNN with pooling.} label{fig:newton_graph}
% \end{figure}

where $\alpha$ is a step length to be chosen and $\epsilon$ is a number to be provided by the user.

First note that there are some obvious numerical parameters that have to be chosen, namely $x_0$ and $\alpha$. However, we note that we can easily make many more. For example we could make $\alpha$ a matrix that depends on the iteration number: $\alpha_n$. Another possibility is to borrow from accelerated gradient methods and other higher order methods and look for an iteration of the form:
\[
    x_{n+1} = \sum_{0 \leq l < K} \beta_{n,l} x_{n-l} + \alpha_{n,l} (F'(x_{n-l}))^{-1}F(x_{n-l}),
\]
where $\beta_{n,l},\alpha_{n,l} \in \mathcal{R}^{d\times d}$. We can go one step further and allow multiple choices per step and choose the best one:
\begin{eqnarray*}
x_{n+1,r} & = & \sum_{0 \leq l < K} \beta_{n,l,r} x_{n-l} + \alpha_{n,l,r} (F'(x_{n-l}))^{-1}F(x_{n-l}), \qquad 1 \leq r \leq N, \\
x_{n+1} &=& \argmin_{x_{n+1,r}} |F(x_{n+1,r})|
\end{eqnarray*}
When we generalize so much it is good to observe that the original trusted Newton method can be recovered for special choices of the new weight matrices $\alpha_{n,l,r}$ and $\beta_{n,l,r}$. This is an important observation as when this heuristic is unrolled and trained via one's favorite machine learning framework~(Figure~\ref{fig:newton_graph}) we are assured of good starting weights with a known performance threshold.

Note also that rather than just setting $x_0=0$ we could also use a more complicated expression like
\[
    x_0 = \sum_{i\in R^d,j} a_{j,i} \gamma_{j,i},
\]
or some more intelligent class of heuristics based on root localization theorems \cite{bini1996numerical}.

\subsection*{A simple heuristic for image classification}

We now present a simple heuristic for the classical MNIST and Fashion MNIST data classification problems.

The MNIST dataset consists of example images of handwritten digits from 0 to 9. The goal is to come up with an algorithm that is capable of recognizing similar types of handwritten digits. As can be seen, there is no precise mathematical formulation of the problem other than what is specifiable via the ground-truth data in the collection. Nevertheless as humans we have some pre-conceived notions of how handwritten digits could be recognized.

For example, we use the common heuristic:
\begin{enumerate}
    \item Choose a representative number of samples from the training data for each digit.
    \item Given a new query image, compare using some appropriate metric the query to the set of representative samples.
    \item Based on the distance to the representative samples, make a decision on which digit the query corresponds to.
\end{enumerate}
The performance of this heuristic depends crucially on the image metric that is used. It is well-known that classical norm-based distance functions tend to perform poorly and many alternatives have been proposed in the literature \cite{simard1993efficient}. Now we describe the heuristic we used for the image metric.

In MNIST the images are all of the same size and shape with the digit roughly in the middle. To compare two images we proceed as follows. Let $X,Y : R^2 \rightarrow R^d$ and
\[
    \sigma(X,Y) = \min_{\theta_1 \leq \theta \leq \theta_2} \|W_1(X - Y \circ R_\theta)\|_p,
\]
where $R_\theta$ denotes rotation by an angle $\theta$, $W_1$ denotes a linear operator, and $\|\cdot\|_p$ denotes the standard $p$-norm. Note that $\sigma$ measures the distance between two images of the same size by considering the minimum over all rotations in the range from $\theta_1$ to $\theta_2$. We are assuming here that the images $X$ and $Y$ have $d$-dimensional pixels.

Next let
\[
\phi(X, Y) = \min_{|i| < w, |j| < h} \sigma(X, Y \circ \tau_{i,j}),
\]
where $\tau_{i,j}$ denotes translation by the vector $(i,j)$. Let $\mu(X,Y)\in \mathcal{R}^2$ be defined such that
\[
\phi(X,Y) = \sigma(X, Y \circ \mu(X,Y)).
\]
Let $\nu(X,i,j,s)$ denote the sub-image of $X$ of size $s \times s$ centered at $(i,j)$ with pixels outside the region set to 0. Then let
\[
    \gamma(X, Y) = \int_{R^2} \phi\left( \nu(X, x, y, s), \nu(Y, x, y, s) \right) w_2(x,y) (1+\|\nabla^2_{x,y}(\mu\left( \nu(X, x, y, s), \nu(Y, x, y, s) \right))\|) \dd x\dd y,
\]
where $w_2$ is a weight function. We take $\gamma(X,Y)$ as our measure of the distance between image $X$ and $Y$. In spite of its messy appearance, the image metric is quite simple: it is computing the distance between the pixels in $X$ and $Y$ by comparing patches of size $s \times s$, and when it compares patches it allows a little bit of translation and rotation, picking the closest match in each case. Then it looks at the induced optical flow and further penalizes those distance where the Laplacian of the flow is large. We believe that this type of heuristic is not uncommon in the literature \cite{yang2007hybrid}.

This heuristic can now be unrolled into a network and more parameters introduced as needed.

\section{Experiments}
In this section, we provide some initial numerical evidence to substantiate our network design methodology on the previously mentioned problems of polynomial root finding and image classification. In particular, we demonstrate how simple heuristics for these problems can be discretized and compiled (currently, manually) into neural network algorithms, and trained for better results using one's favorite machine learning framework. For more detail, we refer readers to the Supplementary Information.

\subsection*{DeepNewtonNet for 1D and 2D polynomial systems}
% A simple realization of the Newton heuristic can be achieved by unrolling the first $n$ Newton iterations, and writing 
% One strategy for implementing the previously described Newton heuristic, is to 
The $n$-iterations of the previously described Newton heuristic can be unrolled and written as a simple $n$-layer network, by computing at each layer:
\begin{align}
    % x_{n+1} &= \sum_{0 \leq l < K} \beta_{n,l} x_{n-l} + \alpha_{n,l} F'(x_{n-l}). \qquad \text{\textbf{NEED TO FIX THIS}}\\
    x_{n+1,r} &= \sum_{0 \leq l < D} \beta_{n,l}^{k_1} x_{n-l} + \gamma_{n,l}^{k_2}F'(x_{n-l},S) - \alpha_{n,l}^{k_3} J^{\dagger}(x_n)F(x_n,S) \nonumber \\
    & \qquad\qquad\qquad\qquad\qquad\qquad \forall \; \beta_{n,l}^k \in B_n, \gamma_{n,l}^k \in C_n, \alpha_{n,l}^k \in A_n \nonumber
    % x_{n+1} &= \argmin_k \| F(q_n^k, S) \|_2^2 \,\qquad \forall \; n
\end{align}
where $S \in \mathcal{R}^c$ represents the coefficients of a polynomial system in some basis (e.g. monomial), \mbox{$F(x,S):\mathcal{R}^d \times \mathcal{R}^c \rightarrow \mathcal{R}^p$} represents the evaluation of each of the $p$ polynomials in this system at the points $x \in \mathcal{R}^d$, and $J^{\dagger}(x,S):\mathcal{R}^d \times \mathcal{R}^c \rightarrow \mathcal{R}^{d \times p}$ represents the generalized inverse of the Jacobian of $F$ with respect to $x$. As mentioned, the parameters of this algorithm are the \textit{tensorized} weights (e.g. matrices) representing possible step-lengths, hysteresis, and momentum factors at each iteration; these possibilities are generated in the network by taking the Cartesian product of the sets $A_n$, $B_n$, and $C_n$. Notice here that $F$, $F'$, and $J^{\dagger}$ are themselves polynomial networks, composed of purely additions and multiplications (e.g. computed via Horner's rule, or similar network).

% \begin{algorithm}[hbt!]
%     \caption{DeepNewton : a tunable $d$-dimensional Newton's Method with Line Search}\label{alg:DeepNewton}
%     \begin{algorithmic}
%         \Procedure{newtonNet}{$S$} \Comment{Given the coefficients of a polynomial system}
%             \State $\hat{x} \gets$ init($S$) \Comment{Simple initial guess (can vary)}
%             \For{$n \in N$}
%                 \State $\hat{x} \gets \argmin_x H$($x_k$,$S$) $ \forall x_k \in $ update($\hat{x}$,$S$)
%             \EndFor
%             \State return $\hat{x}$
%         \EndProcedure
%         \Procedure{newtonUpdate}{$x$,$S$}
%         \State $x \gets \alpha_n^k x + \beta_l^k x_{n-l}$
%         \EndProcedure
%     \end{algorithmic}
% \end{algorithm}

\begin{figure}[hbt!]
    \centering
    \includegraphics[width=\textwidth]{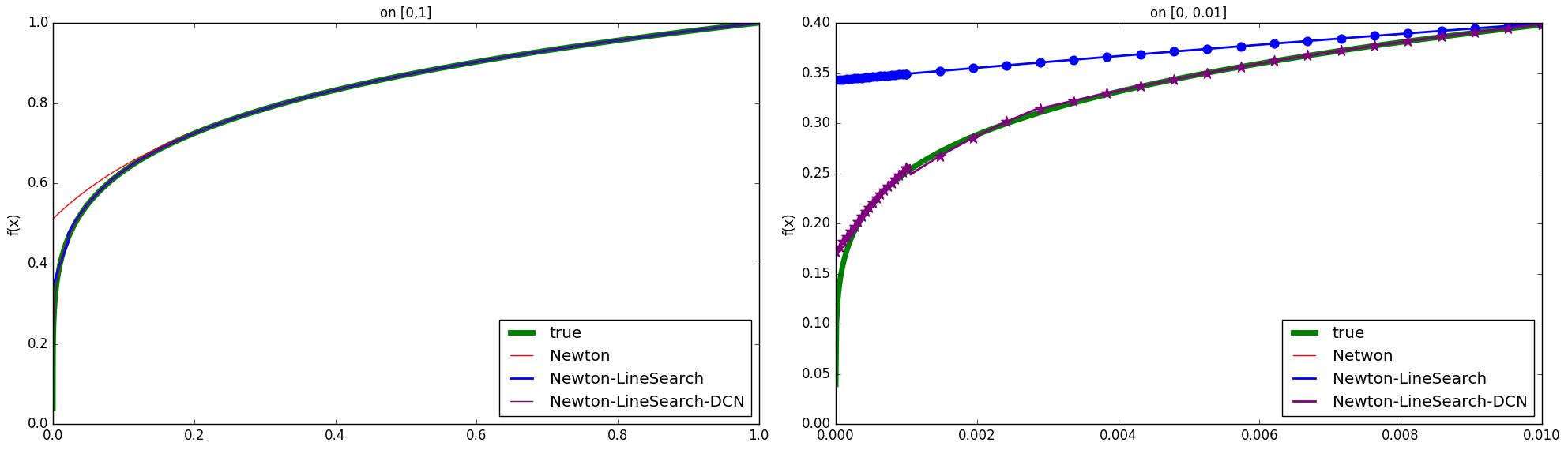}
    \caption{Behavior of the Deep Newton network on a class of simple 1-D polynomials, $x^5 - S = 0$.} %Although the true real-valued root (plotted in green) was not supplied to the network, training the Deep Newton network on the residual error provided significant improvements in approximation performance for this class of polynomial.}
    \label{fig:newton_aroot}
\end{figure}

For simplicity, in our initial tests we fix the degree $D=6$ and number of iterations $n=3$, initialize parameters as \mbox{$A_n=\{ 0.5, 1.0, 1.5 \}\cdot \mathcal{I}_d$}, $B_n=\{\mathcal{I}_d\}$, $C_n=\{0\}$, compute a baseline performance using these weights, and continue training on the residual error (i.e. without ground-truth data). A summary of our initial results on a small class of real 1D and 2D polynomial systems is displayed in Table~\ref{tab:newton}, while the behavior of the method for an even smaller subclass is depicted in Fig.~\ref{fig:newton_aroot}. Note, that in these tests we intentionally picked $x_0$, $S$, and all weights as real numbers so the network would only produce real-valued root estimates (even though imaginary roots exist in the general case)\footnote{We refer readers to the supplementary information for additional notes about these choices, network specifics, and the test metrics that were used in the evaluation.}.
% \begin{figure}[hbt!]
%     \centering
%     % \includegraphics[width=\textwidth]{fig_aroot_fig0alpha5_NewtonIter3.png}
%     \caption{Operation of the Deep Newton network on a simple 1-D and 2-D polynomial system.}
%     \label{fig:newton_polysys}
% \end{figure}

\begin{table}[h!]
    \vspace{-2mm}
    \centering
    \small
    \begin{tabular}{c | c | c | c | c}
      \textit{Method} & \textit{square-root} & \textit{fifth-root} & \textit{1D-poly} & \textit{2D-poly} \\
      \hline
      Newton & 0.7658 & 2.2511 & 0.0588 & - \\
      Newton-LS & 0.0195 & 0.1286 & 0.0699 & 0.0331 \\
      DeepNewton-LS & \textbf{0.0070} & \textbf{0.0560} & \textbf{0.0287} & \textbf{0.0225}\\
    \end{tabular}
    \vspace{2mm}
    \caption{Relative performance (MSE) of selected methods on a small class of polynomial systems.} \label{tab:newton}
  \vspace{-4mm}
\end{table}

% \begin{figure}[hbt!]
%     \centering
%     % \includegraphics[width=\textwidth]{fig_aroot_fig0alpha5_NewtonIter3.png}
%     \caption{Performance of selected methods on small classes of 1-D and 2-D polynomial systems.}
%     \label{fig:newton_polysys}
% \end{figure}

%\newpage
\subsection*{ClusterNet for Image Classification}
% Similarly, a discretized version of the previously described heuristic for image classification can be defined as follows:
Similarly, one possible discretization of the previously described heuristic for the classification of RGB-K images can be realized by computing:
\begin{align*}
    q_k^t &= \|(m_k \odot x) - S_t \hat{x}_k\|_1 * \mathcal{I}_3 \quad \forall \quad k \in K,\quad t\in T \\
    r_k &= \min_t q_k^t \\
    s_k &= \argmin_t q_k^t \\
    d_k &= \|(1 + \lambda \cdot \mathcal{L}\{s_k\}) \odot r_k \|_F^2 \\
    g_k &= \frac{e^{-d_k}}{\sum_j e^{-d_j}} \\
    f &= \sum_k \hat{y}_k \cdot (Q \cdot g)_k %Y \cdot (Q\cdot g)
\end{align*}
where $x \in \mathcal{R}^n$ is the query image, $\hat{x}_k \in \mathcal{R}^n$ and $\hat{y}_k \in \mathcal{R}^m$ are free parameters initialized as selected cluster centers (e.g. random samples from each class) and their label vectors, $m_k$ represents a free weight mask that is initialized conformally to $1$, $\odot$ represents point-wise multiplication (Schur-product), $S_t$ represents a translation-shift operator (with fill-value 0), $*$ represents convolution over the ``valid'' region, $\mathcal{L}$ represents a discrete Laplace operator $\mathcal{L} = \nabla^2$ (e.g. computed using simple 1st-order finite-differences), $\lambda$ represents a free weight on the magnitude of the Laplacian, and we have used the square-Frobenius norm inside the network as a natural replacement for the 2-norm distance for operation on images. Finally, we note the use of a softmax-layer to produce vector $g$, which is used to compute a weighted-average of the cluster labels after multiplication with $Q$ (i.e.~a fully-connected layer with linear activation). $Q$ can be initialized trivially as identity, or with the the eigenvectors of the generalized Gramian matrix of the cluster centers; this later initialization corresponds to an interpretation of the algorithm as a version of spectral clustering, or non-linear PCA, with respect to the heuristic distance function defined by $d_k$ \cite{baldi1989neural,hyvarinen2000independent,cho2009kernel}. 

This algorithm's corresponding signal-flow graph is shown below in Fig.~\ref{fig:cluster_graph}.
\begin{figure}[hbt!]
    \centering
    \includegraphics[width=\textwidth]{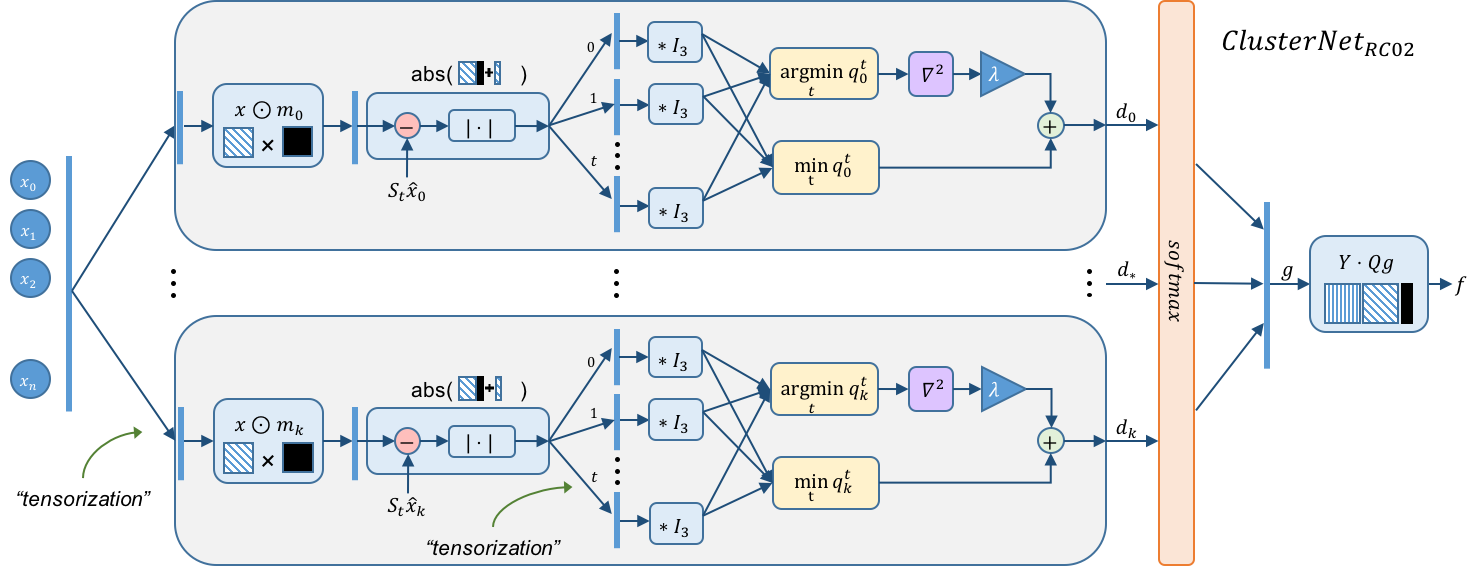}
    \caption{The computation graph of a ``tensorized'' cluster-based classification network (ClusterNet).} \label{fig:cluster_graph}
    \vspace{-3mm}
\end{figure}

We tried various choices for the structural parameters, $\{K,T\}$, which control the width of the network in different ways (e.g. if loops are allowed in the network, then $T$ controls the number of internal loop-iterations), and evaluated the baseline classification performance of these methods on the aforementioned datasets using a sensible initialization for the remaining parameters. Once a reasonable baseline was achieved, we trained the network using stochastic gradient descent and observed strictly better performance. In general, we observed an increase in the baseline training and testing accuracy as the total number of cluster centers increased, but this improvement is somewhat incremental even with training. A thorough analysis of this result will be presented elsewhere, but we comment that this behavior is common in clustering algorithms, and could be alleviated by inclusion of a cluster-repulsion term in the objective as in~\cite{chandrasekaran2017fast}.

The confusion matrices corresponding to ClusterNet performance on the MNIST and Fashion-MNIST testing data is reported in Tables~\ref{tab:mnist1}-\ref{tab:mnist2}~and~\ref{tab:fmnist1}-\ref{tab:fmnist2} respectively. Each network was initialized with a reasonable number of cluster centers (10 and 25 per class, respectively), relative to the number of parameters used in other state of the art solutions \cite{rawat2017deep}, and trained for sufficient time to show improvement ($\sim10$ epochs for MNIST, $\sim100$ epochs for Fashion-MNIST). The point here is not to show that our networks are more ``optimal'' than conventional networks; rather, we demonstrate that well-loved heuristics can (in some cases) perform as well as DNNs when their graphs are sufficiently parametrized and, most importantly, are allowed to train with respect to a corpus of data.

\vspace{-3mm}
% \begin{wraptable}{c}{\textwidth}
\begin{table}[hbt!]
    \caption{Test-Performance of a ClusterNet implementation on MNIST-Digits (10x10~cluster centers).} \label{tab:imageclass}
    \begin{subtable}{0.5\linewidth}
        \centering
        \scriptsize
        \tabcolsep=0.10cm
        \begin{tabular}{c|c c c c c c c c c c |}
            & \multicolumn{10}{c}{Predicted Label} \\
            & 0 & 1 & 2 & 3 & 4 & 5 & 6 & 7 & 8 & 9 \\ 
          \hline
            0 & \textbf{85.3} & 0.0 & 0.5 & 0.4 & 0.0 & 5.3 & 6.0 & 0.0 & 2.4 & 0.0 \\
            1 & 0.0 & \textbf{95.8} & 3.2 & 0.3 & 0.2 & 0.1 & 0.2 & 0.2 & 0.0 & 0.0 \\
            2 & 1.7 & 0.6 & \textbf{86.4} & 2.5 & 0.6 & 0.4 & 4.2 & 1.4 & 2.2 & 0.0 \\
            3 & 0.0 & 1.3 & 2.6 & \textbf{78.0} & 0.8 & 12.9 & 1.1 & 1.1 & 1.8 & 0.4 \\
            4 & 0.1 & 2.0 & 0.8 & 0.1 & \textbf{66.1} & 0.5 & 8.9 & 4.2 & 2.2 & 15.1 \\
            5 & 1.3 & 2.0 & 1.5 & 8.4 & 1.9 & \textbf{75.3} & 6.6 & 0.2 & 1.6 & 1.1 \\
            6 & 1.6 & 1.6 & 1.0 & 0.7 & 0.3 & 3.0 & \textbf{91.6} & 0.0 & 0.1 & 0.0 \\
            7 & 0.5 & 3.7 & 4.4 & 0.4 & 8.7 & 2.3 & 0.1 & \textbf{68.2} & 1.6 & 10.1 \\
            8 & 0.4 & 4.2 & 3.4 & 11.7 & 1.6 & 6.9 & 3.3 & 1.8 & \textbf{66.1} & 0.5 \\
            9 & 1.2 & 1.5 & 0.3 & 1.2 & 19.8 & 0.5 & 1.5 & 10.4 & 2.8 & \textbf{60.8}
        \end{tabular}
        % \vspace{3mm}
        \subcaption{Pre-training, 77.4\% overall.} \label{tab:mnist1}
        % \vspace{-2mm}
  \end{subtable}
  \begin{subtable}{0.5\linewidth}
        \centering
        \scriptsize
        \tabcolsep=0.10cm
        \begin{tabular}{c| c c c c c c c c c c |}
            & \multicolumn{10}{c}{Predicted Label} \\
            & 0 & 1 & 2 & 3 & 4 & 5 & 6 & 7 & 8 & 9 \\ 
          \hline
            0 & \textbf{98.6} & 0.1 & 0.1 & 0.1 & 0.1 & 0.2 & 0.2 & 0.1 & 0.3 & 0.2 \\
            1 & 0.0 & \textbf{98.8} & 0.4 & 0.1 & 0.0 & 0.0 & 0.2 & 0.1 & 0.5 & 0.0 \\
            2 & 0.2 & 0.0 & \textbf{96.7} & 0.5 & 0.3 & 0.1 & 0.4 & 1.0 & 0.9 & 0.0 \\
            3 & 0.0 & 0.0 & 0.3 & \textbf{96.6} & 0.1 & 1.2 & 0.0 & 0.6 & 0.5 & 0.7 \\
            4 & 0.0 & 0.1 & 0.5 & 0.0 & \textbf{97.0} & 0.1 & 0.4 & 0.4 & 0.1 & 1.3 \\
            5 & 0.3 & 0.1 & 0.0 & 1.1 & 0.3 & \textbf{95.7} & 0.9 & 0.3 & 0.7 & 0.4 \\
            6 & 0.4 & 0.2 & 0.4 & 0.1 & 0.4 & 0.8 & \textbf{97.0} & 0.0 & 0.6 & 0.0 \\
            7 & 0.0 & 0.3 & 0.9 & 0.3 & 0.2 & 0.0 & 0.0 & \textbf{97.2} & 0.4 & 0.8 \\
            8 & 0.4 & 0.0 & 0.2 & 1.0 & 0.2 & 0.9 & 0.1 & 0.3 & \textbf{96.0} & 0.8 \\
            9 & 0.2 & 0.3 & 0.0 & 0.4 & 0.8 & 0.2 & 0.0 & 1.0 & 0.3 & \textbf{96.8}
        \end{tabular}
        % \vspace{3mm}
        \subcaption{Post-training, 97.1\% overall.} \label{tab:mnist2}
        % \vspace{-2mm}
  \end{subtable}
  \vspace{-3mm}
\end{table}

\begin{table}[hbt!]
    \caption{Test-Performance of a ClusterNet implementation on Fashion-MNIST (25x10 and 10x10~cluster centers respectively).}
    \begin{subtable}{0.5\linewidth}
        \centering
        \scriptsize
        \tabcolsep=0.11cm
        \begin{tabular}{c| c c c c c c c c c c |}
            & \multicolumn{10}{c}{Predicted Label} \\
            & 0 & 1 & 2 & 3 & 4 & 5 & 6 & 7 & 8 & 9 \\ 
          \hline
            0 & \textbf{77.1} & 1.5 & 3.1 & 6.6 & 1.6 & 0.4 & 8.4 & 0.0 & 1.3 & 0.0 \\
            1 & 1.2 & \textbf{92.7} & 0.5 & 4.2 & 0.7 & 0.1 & 0.5 & 0.0 & 0.1 & 0.0 \\
            2 & 2.2 & 1.4 & \textbf{54.5} & 0.6 & 19.8 & 0.2 & 20.6 & 0.0 & 0.7 & 0.0 \\
            3 & 8.8 & 5.0 & 1.1 & \textbf{73.6} & 6.6 & 0.1 & 3.5 & 0.0 & 1.3 & 0.0 \\
            4 & 0.7 & 1.4 & 19.0 & 4.3 & \textbf{58.1} & 0.2 & 15.4 & 0.0 & 0.9 & 0.0 \\
            5 & 0.0 & 0.0 & 0.0 & 0.1 & 0.0 & \textbf{88.3} & 0.3 & 6.8 & 1.0 & 3.5 \\
            6 & 23.5 & 1.2 & 17.9 & 3.5 & 14.0 & 1.0 & \textbf{37.2} & 0.0 & 1.7 & 0.0 \\
            7 & 0.0 & 0.0 & 0.0 & 0.0 & 0.0 & 8.7 & 0.0 & \textbf{82.0} & 0.3 & 9.0 \\
            8 & 0.6 & 0.1 & 3.6 & 0.6 & 1.3 & 2.9 & 1.8 & 0.4 & \textbf{88.6} & 0.1 \\
            9 & 0.0 & 0.0 & 0.0 & 0.0 & 0.0 & 3.6 & 0.0 & 6.7 & 0.1 & \textbf{89.6}
        \end{tabular}
        % \vspace{3mm}
        \subcaption{Pre-training (25x10 clusters), 74.1\% overall.} \label{tab:fmnist1}
        % \vspace{-2mm}
  \end{subtable}
  \begin{subtable}{0.5\linewidth}
        \centering
        \scriptsize
        \tabcolsep=0.11cm
        \begin{tabular}{c| c c c c c c c c c c |}
            & \multicolumn{10}{c}{Predicted Label} \\
            & 0 & 1 & 2 & 3 & 4 & 5 & 6 & 7 & 8 & 9\\ 
          \hline
            0 & \textbf{85.4} & 0.0 & 1.9 & 0.9 & 0.5 & 0.1 & 10.2 & 0.0 & 1.0 & 0.0 \\
            1 & 0.4 & \textbf{97.4} & 0.4 & 0.9 & 0.4 & 0.0 & 0.4 & 0.0 & 0.1 & 0.0 \\
            2 & 1.2 & 0.2 & \textbf{83.9} & 0.7 & 7.4 & 0.0 & 6.3 & 0.0 & 0.3 & 0.0 \\
            3 & 2.8 & 0.9 & 1.2 & \textbf{89.2} & 2.6 & 0.0 & 2.9 & 0.0 & 0.4 & 0.0 \\
            4 & 0.0 & 0.0 & 6.8 & 2.8 & \textbf{84.7} & 0.0 & 5.4 & 0.0 & 0.3 & 0.0 \\
            5 & 0.0 & 0.0 & 0.0 & 0.1 & 0.0 & \textbf{97.7} & 0.0 & 1.4 & 0.1 & 0.7 \\
            6 & 10.1 & 0.3 & 7.2 & 2.4 & 7.2 & 0.0 & \textbf{71.0} & 0.0 & 1.8 & 0.0 \\
            7 & 0.0 & 0.0 & 0.0 & 0.0 & 0.0 & 1.2 & 0.0 & \textbf{96.7} & 0.0 & 2.1 \\
            8 & 0.4 & 0.0 & 0.4 & 0.2 & 0.2 & 0.1 & 0.4 & 0.3 & \textbf{98.0} & 0.0 \\
            9 & 0.0 & 0.0 & 0.0 & 0.0 & 0.0 & 0.8 & 0.1 & 3.0 & 0.0 & \textbf{96.1}
        \end{tabular}
        % \vspace{3mm}
        \subcaption{Post-training (10x10 clusters), 90.01\% overall.} \label{tab:fmnist2}
  \end{subtable}
%   \vspace{-2mm}
%   \begin{subtable}{\linewidth}
%         \centering
%         % \begin{center}
%         \begin{tabular}{c| c c c c c c c c c c | c}
%             & \multicolumn{10}{c}{Predicted Label} \\
%             & 0 & 1 & 2 & 3 & 4 & 5 & 6 & 7 & 8 & 9 & \\ 
%           \hline
%             0 & 39.5 & 6.4 & 9.0 & 2.4 & 5.8 & 5.8 & 6.6 & 3.9 & 12.5 & 8.2 & 39.5\% \\
%             1 & 11.6 & 19.5 & 2.5 & 10.0 & 9.9 & 14.2 & 8.2 & 19.5 & 1.8 & 2.8 & 19.5\% \\
%             2 & 15.0 & 4.9 & 23.2 & 3.8 & 3.2 & 3.2 & 6.1 & 4.6 & 16.5 & 19.4 & 23.2\% \\
%             3 & 10.6 & 15.6 & 4.8 & 10.5 & 10.0 & 12.9 & 13.4 & 14.2 & 2.0 & 6.0 & 10.5\% \\
%             4 & 11.5 & 19.9 & 3.8 & 10.0 & 18.1 & 9.0 & 11.9 & 11.2 & 2.5 & 2.1 & 18.1\% \\
%             5 & 7.9 & 16.0 & 4.0 & 10.9 & 8.2 & 23.4 & 11.9 & 14.4 & 1.1 & 2.2 & 23.4\% \\
%             6 & 5.0 & 9.1 & 1.9 & 8.0 & 7.6 & 20.2 & 28.5 & 14.4 & 2.0 & 3.2 & 28.5\% \\
%             7 & 6.0 & 12.9 & 2.5 & 11.2 & 6.6 & 16.1 & 16.1 & 22.9 & 2.2 & 3.4 & 22.9\% \\
%             8 & 11.5 & 2.5 & 8.4 & 1.6 & 1.2 & 2.0 & 3.0 & 3.2 & 49.9 & 16.6 & 49.9\% \\
%             9 & 9.1 & 3.5 & 14.4 & 4.0 & 2.8 & 2.8 & 5.9 & 6.9 & 25.4 & 25.4 & 25.4\% \\
%         \end{tabular}
%         \vspace{3mm}
%         \caption{STL-10 Test-Confusion Matrix (Init, pre-training with 25x10 cluster centers), 26.09\% overall.} \label{tab:stl10} %trainacc : 0.2948
%         % \end{center}
%         \vspace{-2mm}
%   \end{subtable}
\vspace{-6mm}
\end{table}
% \end{wraptable}

As seen in Table~\ref{tab:mnist1}, our heuristic performs quite well on the MNIST data ($77.4\%$ test accuracy) initialized with just 10 examples from each class, and no other training. However, even with a modest amount of training (Table~\ref{tab:mnist2}), we are able to match the performance of DNN approaches ($\sim97\%+$).
% In general, the Fashion-MNIST dataset proved a little more difficult for the heuristic initially.
To achieve around the same level of initial accuracy in the Fashion-MNIST dataset, around 20-30 examples were needed from each class~(Table~\ref{tab:fmnist1}); however, the testing accuracy in these cases would sometimes saturate after training for $100$-epochs to values not substantially better than the $\sim90\%+$~result that was achieved after training with just 10 examples from each class (Table~\ref{tab:fmnist2}). Since the purpose of this paper is not to demonstrate an effective training strategy for the proposed style of algorithmic networks, but instead to show that these networks are viable in terms of capacity, we have only depicted the results for the 10-example-per-class case, which was achieved using only naive optimization strategies (e.g. stochastic gradient descent on a mean-square error residual). We believe further experimentation using special training algorithms and longer training periods can produce even better results.

% \newpage
% \subsection*{KernelNet for Data Synthesis}

% \clearpage
% \newpage
\section{Discussion}
We do not claim that the algorithms and networks presented in this paper are optimal for solving the presented problems.  While in this case our networks achieved a desirable performance, it is possible that parametrizations based on human intuition of the problem can make the task of finding the optimal weights more challenging than is necessary, though this might be compensated by providing better initial weights \cite{mishkin2015all}. With that being said, we note that our DeepNewton and ClusterNet implementations did \textbf{not} require any sigmoidal or ReLU-like activation functions in the network, and instead rely on a purely ``polynomial'' implementation composed of adds and multiplies.

Moreover by embedding the heuristic algorithmic networks into larger DNNs, we have a methodical strategy for picking initial weights with a known performance threshold. Once embedded into a larger network, the weights can be tuned using standard techniques, such as back propagation. In a sense, this can be thought of as transfer learning, where the model weights and architecture are adopted from a heuristic that the algorithm designer believes would perform reasonably, even if it may not be optimal, for the given task.

In general, we believe that our design methodology will extend well to heuristics used in other computer vision tasks. For example, in video processing and understanding, it was common for motion features to be hand-crafted. When viewed as a heuristic, these features can be naturally initialized, generalized, and trained in an end-to-end algorithm. We believe this viewpoint will help practitioners re-consider leveraging algorithms (heuristics), that historically have shown promise but have been largely discarded in the wake of deep learning. This will also aid in the design of networks with \textsl{understandable} and \textsl{predictable} performance, and also with better composability properties.

We note that we do not address the problem of algorithm (or network) synthesis or concerns related with differentiable interpreters.

\small
\bibliographystyle{unsrt}
\bibliography{nips2018_DeepAlgos}

\end{document}